%% file: templateArxiv.tex
\documentclass{article}

\usepackage{PRIMEarxiv}

\usepackage[utf8]{inputenc} 
\usepackage[T1]{fontenc}    
\usepackage{hyperref}       
\usepackage{url}            
\usepackage{booktabs}       
\usepackage{amsfonts}       
\usepackage{nicefrac}       
\usepackage{microtype}      
\usepackage{lipsum}
\usepackage{fancyhdr}       
\usepackage{graphicx}       
\graphicspath{{media/}}     

\title{STEMS: Spatial-Temporal Mapping For \\Spiking Neural Networks
\thanks{Code is available at \url{https://github.com/sjoks93/STEMS/}} 
}

\author{
  Sherif Eissa  \\
  Eindhoven University \\
  of Technology \\
  \texttt{s.s.b.eissa@tue.nl} \\
   \And
  Sander Stuijk \\
  Eindhoven University \\
  of Technology \\
  \texttt{s.stuijk@tue.nl} \\
   \And
  Floran de Putter \\
  Eindhoven University \\
  of Technology \\
  \texttt{f.a.m.d.putter@email} \\  
   \And
  Andrea Nardi-Dei \\
  Eindhoven University \\
  of Technology \\
  \texttt{a.y.f.nardi.dei.da.filicaia.dotti@tue.nl} \\  
   \And
  Federico Corradi \\
  Eindhoven University \\
  of Technology \\
  \texttt{f.corradi@tue.nl} \\
   \And
  Henk Corporaal \\
  Eindhoven University \\
  of Technology \\
  \texttt{h.corporaal@tue.nl} \\ 
}

\begin{document}
\maketitle

\begin{abstract}
Spiking Neural Networks (SNNs) are promising bio-inspired third-generation neural networks. Recent research has trained deep SNN models with accuracy on par with Artificial Neural Networks (ANNs). Although the event-driven and sparse nature of SNNs show potential for more energy efficient computation than ANNs, SNN neurons have internal states which evolve over time. Keeping track of SNN states can significantly increase data movement and storage requirements, potentially losing its advantages with respect to ANNs. This paper investigates the energy effects of having neuron states, and how it is influenced by the chosen mapping to realistic hardware architectures with advanced memory hierarchies.

Therefore, we develop STEMS, a mapping design space exploration tool for SNNs. STEMS models SNN’s stateful behavior and explores intralayer and interlayer mapping optimizations to minimize data movement, considering both spatial and temporal SNN dimensions. Using STEMS, we show up to 12x reduction in off-chip data movement and 5x reduction in energy (on top of intra-layer optimizations), on two event-based vision SNN benchmarks. Finally, neuron states may not be needed for all SNN layers. By optimizing neuron states for one of our benchmarks, we show 20x reduction in neuron states and 1.4x better performance without accuracy loss.
\end{abstract}


\input{introduction}
\input{relatedwork}
\input{background}

\input{methods}
\input{setup}

\input{experiments}

\input{discussion}

\input{conclusion}

\bibliographystyle{unsrt}  
\bibliography{ref_control.bib,references.bib}

\input{appendix}

\end{document}

%% file: introduction.tex
\section{Introduction}

Spiking Neural Networks (SNNs) are biologically inspired neural networks that have gained increasing popularity as a solution for efficient edge AI, due to their sparse spike-based computation. SNNs use state-full neuron models that evolve over time akin to recurrent neural networks, producing sparse spikes over time \cite{8891809}. Direct SNN training has seen recent success, with accuracy on par with other deep learning solutions \cite{sewresnet, 10376840}. Due to their state-full event-driven nature, SNNs are promising candidates for (spatial-)temporal problems such as event-based vision using event-based cameras (i.e Dynamic Vision Sensors (DVS)), which has recently gained popularity as an efficient solution for embedded computer vision~\cite{gallego2020event}.

While SNNs might have better compute efficiency compared to Artificial Neural Networks (ANNs), the performance of deep models is usually dominated by off-chip data movement due to the large amount of memory (weights and neuron states) and features (spikes) that exceed on-chip capacity. Hence, limiting off-chip data movement can be crucial for SNN energy efficiency on memory-constrained devices. Additionally, recent digital neuromorphic architectures consist of shared memory hierarchies and parallel Processing Elements (PEs)~\cite{10132492, 9855834}, which provide opportunities for data-reuse and parallel compute, improving their performance on large SNN models compared to earlier architectures that lack shared memory hierarchies and have poor data reuse \cite{senapati2022thorneuromorphicprocessor}.

Mapping SNNs on an embedded architecture with limited on-chip memory can be challenging. SNNs have recurrent hidden states (i.e neuron states), which need to be remembered and stored throughout SNN inference. Storing neuron states off-chip results in less favorable performance, as they are continuously moved on-chip, updated, and stored back. Mapping design space exploration (DSE) tools have recently been developed to explore the mapping space of image processing kernels to tensor computing architectures, to minimize data movement and improve performance with loop nest optimization, using search heuristics and analytical cost models~\cite{zigzag,9923807}. Other tools also explore the inter-layer mapping space and apply (spatial) layer fusion to reduce (off-chip) data movement by orders of magnitude on memory-constrained devices\cite{stream, Sioutas2020ScheduleSF, 9646923, 8667835}. Inspired by such tools\cite{zigzag, stream}, 
we explore inter-layer mapping optimizations
for SNNs. This has resulted in the following contributions:

 \begin{itemize}
     \item STEMS: Spatial temporal mapping exploration tool for SNNs. STEMS applies intra- and inter-layer mapping optimizations in space and time. 
     \item By applying STEMS's inter-layer DSE on two event-based vision SNN benchmarks, we minimize the effect of neuron states on data movement and demonstrate up to 12x reduction in off-chip data movement and up to 5x reduction in energy. 
     \item With lessons learned from recent event-based vision models \cite{red,10204090,9749022}, we explore neuron state optimization for one of our benchmarks. The results show a 20x reduction in neuron states without accuracy loss and a 1.4x reduction in energy using STEMS. 
 \end{itemize}

 This article is organized as follows: Sections \ref{sec:background} and \ref{sec:related_work} provide background on SNNs and inter-layer mapping, and a brief literature review on SNN mapping. Section \ref{sec:methods} explains STEMS, our mapping space, and the neuron state memory optimization method, followed by our experimental setup and results in Sections \ref{sec:expsetup} and \ref{sec:experiments}. Finally, the article ends with a discussion, conclusions, and recommendations for future work.

%% file: relatedwork.tex
\section{Related work} \label{sec:related_work}

Earlier work on SNN mapping assumed fully distributed multicore architectures with no shared memory, where the focus was on partitioning and mapping neurons on a homogeneous multicore architecture, which is a well-known NP-hard problem \cite{huynh2022implementing}. Except for mapping tools that target specific hardware platforms \cite{10.1145/3192366.3192371}, the optimization goal is to maximize resource utilization while minimizing inter-core communication. This problem has been solved with many different heuristics, such as particle swarm optimization \cite{8342201, 10.1145/3386263.3406900}, graph partitioning \cite{titirsha2020thermal, 9996702}, and integer linear programming \cite{das2022realtime}. Some also consider hardware characteristics such as reliability and endurance \cite{titirsha2021endurance}. All these methods do not consider off-chip memories, so they can not handle situations when an SNN does not fit on-chip. They also do not consider shared on-chip memory resources or loop nest optimizations.

NeuProMa \cite{neuproma} is an SNN mapping framework that considers off-chip data movement, where an SNN can not fit onto on-chip and resources have to be time-multiplexed. Their framework maps SNNs in three steps: split, partition, and map. First, the SNN is divided into several subnetworks that fit on-chip resources. SNNs are split according to one of three splitting strategies: channel split, pixel split, or link split. The link split, which creates depth-first subnets similar to layer fusion, performed overall better than the other two on different benchmarks. Then, each subnetwork is mapped using heuristics to reduce inter-core spike communication. This work considers limited splitting strategies and does not consider shared on-chip memories and loop nest optimizations.

SMART is a design flow for mapping SNNs in resource-constrained heterogeneous neuromorphic systems coupled with a CPU \cite{das2022realtime}. This work breaks down the mapping problem into four smaller sub-problems; throughput lower bound, operation mapping, activation/weight mapping, and parallel scheduling. The lower throughput bound is estimated using a self-timed execution schedule. Operation and activation/weight mapping are done using integer linear programming. For operation mapping, latency and memory requirements are combined in a cost function to determine whether operations (i.e. layers) are to be mapped on neuromorphic cores or a CPU. For activation/weight mapping, scratchpad memories are split into a pinned space and an immediate space. Activation/weights mapped to pinned space are never ejected to main memory, allowing better data reuse, while those mapped to immediate space can be ejected to free up space whenever necessary, with the objective of minimizing data movement between main memory and distributed memories. Finally, a parallel schedule is formed, with task- and batch-level parallelism possible. While this work optimizes data movement, it does not consider intra-layer nor inter-layer optimizations.


In \cite{10.1145/3531437.3539704}, the authors take a similar approach to our work, where they use Timeloop \cite{9923807} to explore efficient data flows for SNN workloads on an Eyeriss-like hardware architecture\cite{7738524}. They explore a set of mapping rules and dataflows. However, they do not explore inter-layer optimizations. 

To our knowledge, our work is the first to explore SNN mapping in neuromorphic architectures, with intra-layer and inter-layer mapping optimization across spatial (layer fusion) and temporal (time batching) dimensions to improve hardware utilization and minimize data movement. We highlight in this work how inter-layer (on top of intra-layer) optimizations can significantly reduce off-chip data movement under stringent on-chip memory capacity, by improving reuse of neuron states over time and reducing the size of intermediate features/spikes.

%% file: background.tex
\section{Background}\label{sec:background}

This section introduces the SNN algorithm commonly used in vision applications and the general behavior of SNNs, as well as inter-layer mapping (i.e., layer fusion). 

\subsection{Spiking Neural Network Algorithm}

Recent advances have enabled direct training of large SNNs for complex event-based vision applications such as object detection and recognition~\cite{sewresnet, 10376840}. SNNs have inter-layer connections (i.e. projections) similar to ANNs such as (strided) spatial convolution layers, batch normalization, residual connections, and spatial pooling. However, there are two key differences between SNNs and ANNs. SNNs operate using sparse spike-based features and have self-recurrent activation functions inspired by biological neurons, whereas ANNs operate on real-valued features and have nonrecurrent (i.e., memoryless) activation functions. SNN neuron activation function is self-recurrent because it depends on the neuron's internal state~\cite{8891809}. Additionally, SNN inference happens over time, while ANN inference is not necessarily over time.  

\begin{figure}[!t]
\centering
\includegraphics[width=.75\columnwidth]{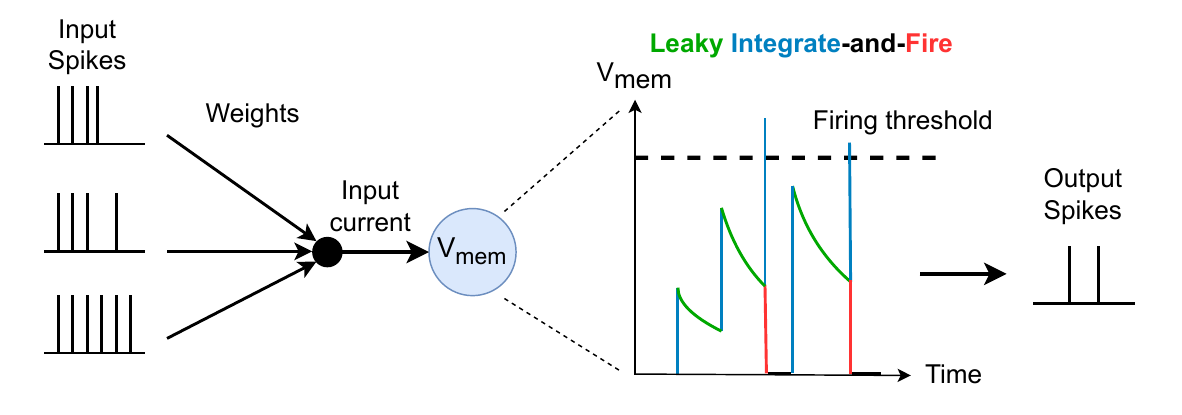}
\caption{Leaky Integrate-and-Fire (LIF) neuron. Colors highlight its 3 features: Leakage, integration, and fire-and-reset. The membrane potential ($V_{mem}$), which is its only state, accumulates input current and controls firing.}
\label{fig:lif}
\end{figure}

SNNs commonly implement the leaky integrate-and-fire (LIF) neuron model, especially for vision applications.This is due to its simplicity in implementation in analog and digital systems, as well as its ability to replicate basic features of biological neurons such as leakage, integration of spikes, and all-or-none action potential generation~\cite{gerstner2014neuronal}. The LIF neuron model has one recurrent state, its membrane potential ($V_{mem}$). Figure \ref{fig:lif} illustrates the dynamics of an LIF neuron. The input current is integrated into the neuron's membrane potential, and the neuron fires and resets when it exceeds its firing threshold, whereas the membrane potential leaks over time.

SNN neurons are simulated digitally using a discrete timestep model, with inferences taking a specific number of timesteps according to the SNN model and initializing neuron states at the beginning of inference.
The LIF discrete timestep model can be simplified to

\begin{equation}
    I[t] = \sum_{proj}{W S_{in}[t]}
\end{equation}
\begin{equation}
V_{mem}[t] = \alpha V_{mem}[t-1] + I[t] - S_{out}[t-1] I_{reset}
\end{equation}
 \begin{equation}
S_{out}[t] = V_{mem}[t] \geq V_{thr} 
 \end{equation}

where $t$ is the discrete timestep, $V_{mem}$ is the membrane potential, $\alpha$ is the membrane leakage, $S_{in}$ and $S_{out}$ are the input and output spikes, respectively, $I$ is the input current, and $V_{thr}$ and $I_{reset}$ are the firing threshold and reset current, respectively. An SNN is computed by traversing these equations across space (i.e., neurons, layers) and time.

\begin{figure}[!t]
\centering
\includegraphics[width=.65\columnwidth]{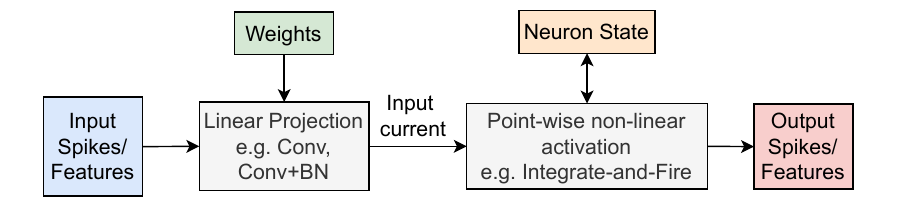}
\caption{Computation graph of one SNN layer during one discrete timestep. The neuron states are used and updated during activation.}
\label{fig:state}
\end{figure}

Figure \ref{fig:state} illustrates the general behavior of an SNN layer during a discrete timestep: \textbf{1.} the input spikes are integrated and modulated by weights (i.e. synapses); \textbf{2.} the input current and output spike are used to update the neuron state, including its membrane potential; \textbf{3.} if the membrane potential exceeds the firing threshold state, it fires. Neuron models can have multiple hidden states (for example, multicompartment models, adaptive models\cite{gerstner2014neuronal}).


\subsection{Layer Fusion in ANNs}
\label{sec:lf}

ANN workloads consist of cascaded blocks, where each block is a directed acyclic graph of layer operators such as spatial convolution, addition (i.e. residual connection), and spatial reductions. Intra-layer loop nest optimizations can improve data reuse and reduce data movement within an operator \cite{zigzag}. However, off-loading and reloading features between operators may be detrimental to performance, which is dominated by off-chip data movement. Hence, applying inter-layer optimizations by considering all operators simultaneously can provide better mapping solutions.

\begin{figure*}[!t]
\centering
\includegraphics[width=\textwidth]{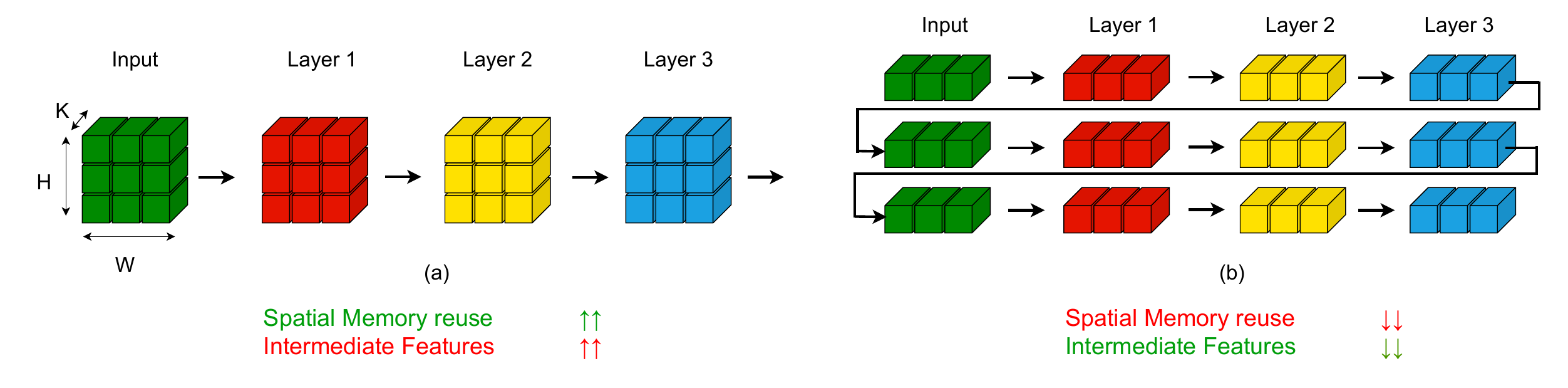}
\caption{Illustration of (a) Layer-by-layer (LBL) schedule versus (b) layer-fused schedule for a 3-layer ANN block with 1D convolution connection (for simplicity). Each layer has output dimensions of size K x H x W, where K is number of channel and H and W are the height and width respectively. Layer-by-layer schedule can maximize reuse of weights, while layer-fused schedule reduces the size of intermediate features.} 
\label{fig:dnn-sch}
\end{figure*}

The typical ANN inter-layer mapping is layer-by-layer (LBL), where one layer is completely scheduled before successive layers can start. In LBL schedules, complete inter-layer features are cached on-chip, if possible, or off-chip. On the other hand, layer-fused (LF) schedules, where layers are partitioned into nodes and interleaved across spatial dimensions (typically line-by-line) can significantly improve performance by reducing the size of cached intermediate features ~\cite{stream, Sioutas2020ScheduleSF, 9646923, 8667835}. This comes at the cost of reducing intra-layer weight reuse across space (\textit{spatial memory reuse}) and increasing the variable lifetime of the weights, which increases the storage of cached weights. Figure \ref{fig:dnn-sch} illustrates LBL and LF schedules on a simple 3-layer ANN block with 1D convolution kernels.




Layer fusion is complementary to input batching. In input batching, multiple input frames are consumed simultaneously, resulting in larger intermediate features, but better reuse of weights (memory). In vision classification models, early blocks have shallow channel dimensions and large spatial dimensions. The channel dimensions grow deeper in deeper layers and the spatial dimensions shrink due to spatial pooling. Hence, earlier blocks tend to have more features, while later blocks tend to have more memory (weights). Hence, earlier blocks typically prefer layer fusion to reduce intermediate feature sizes, while later blocks typically prefer LBL schedules and input-batching to maximize memory (weights) reuse.

%% file: methods.tex
\section{Methods}
\label{sec:methods}
We present an overview on STEMS and its stages, as well as the inter-layer mapping space explored with STEMS. We also motivate and explain our neuron state reduction exploration for spatial (temporal) tasks such as event-based vision.



\subsection{STEMS overview}
\label{sec:ahm}


STEMS is a mapping design space exploration tool for SNNs based on Stream, a state-of-the-art tool for multi-core ANN mapping \cite{stream}. STEMS explores both intra-layer mapping with loop nest optimizations, and inter-layer mapping with workload partitioning and scheduling along spatial (i.e., layer fusion) and temporal dimensions (i.e. time batching). Compared to Stream, STEMS supports temporal and stateful workloads such as SNNs and enables their inter-layer mapping exploration along spatial and temporal dimensions.

\begin{figure}[t]
    \centering
    \includegraphics[width=.45\columnwidth]{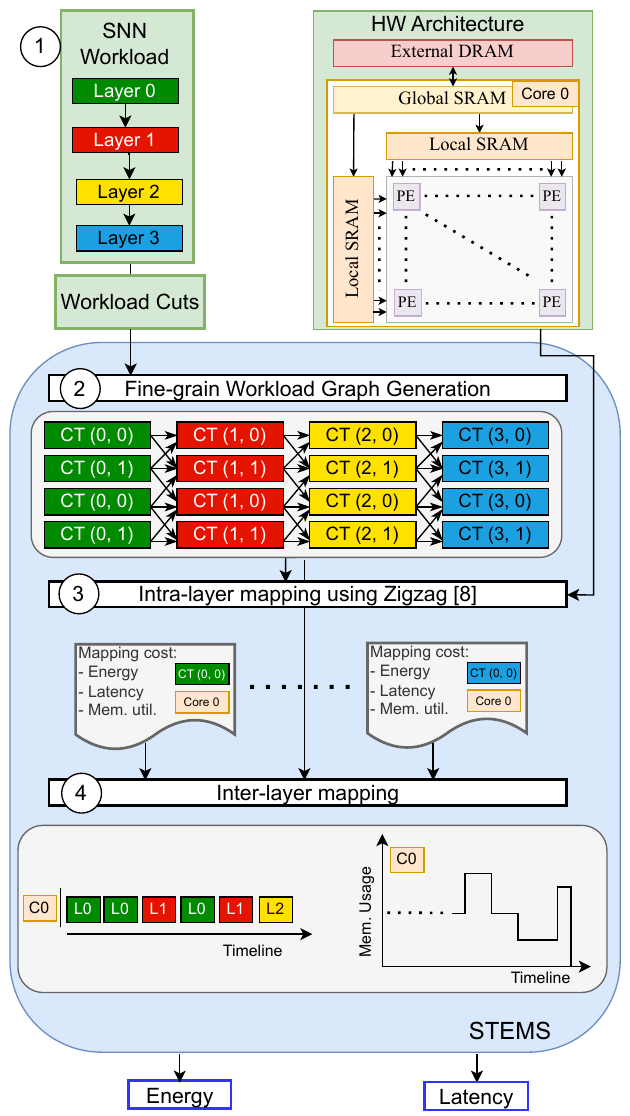}
    \caption{Overview of the STEMS framework.}

    \label{fig:stems}
\end{figure}

Figure \ref{fig:stems} shows an overview of STEMS. STEMS breaks down the mapping problem into pipeline stages similar to Stream. STEMS has three user-defined inputs: the workload and accelerator descriptions, and user-defined workload cuts along spatial and temporal dimensions. STEMS pipeline can be described in the following four stages: \textbf{1.} user input is parsed, and the workload and accelerator models are generated; \textbf{2.} fine-grain workload graph is generated according to user-defined workload cuts; \textbf{3.} intra-layer mapping is generated by optimizing each unique tile-core mapping; \textbf{4.} inter-layer schedule is generated by scheduling the fine-grained workload graph in a specific order while managing hardware resources. For multi-core accelerators, a genetic algorithm can be used in this stage to explore layer-to-core assignments.

\subsection{STEMS components}

\subsubsection{Input parsing}

In this stage, the workload and accelerator models are generated from user input. 

The accelerator model consists of an external memory and one or more computation cores connected in a NoC. The NoC has a mesh structure and is defined by the data bandwidth and data movement costs between cores and between any core and external memory. Only one core can request data from the external memory at a time. Each core is described by a memory hierarchy that can store operand tiles at different memory levels, and an array of parallel processing elements. The memory blocks in the hierarchy are defined by their size, bandwidth (bits per cycle), and energy (per bit). While the processing elements are defined by their energy per operation, size, and unrolled dimensions.

The workload is defined as an acyclic graph of operators. Each operator consists of a set of operands (e.g Weights, Neuron States, Features) and a set of dimensions (e.g Time, Space, Channels) and a set of linear projections defined by the relationship between operands (e.g Convolution). 

To support SNNs and temporal workloads, neuron states and time are incorporated as additional workload features. Two types of neuron state can be supported: the accumulator state (i.e. membrane potential of LIF neurons) and other states used during neuron activation (i.e. threshold adaptation, multi-compartment neuron states). Each operator has a block id that defines the granularity of user-defined workload cuts.


\subsubsection{Fine-grain workload graph generation}

In this stage, a fine-grained workload graph is generated out of the workload graph and user-defined workload cuts. Such cuts define how the operators in each block are partitioned into smaller \textit{computation tiles} (CT) that are used in intra- and inter- layer mapping stages. Such partitioning also defines how these computation tiles will be scheduled in the inter-layer mapping stage. 

To support SNNs and temporal workloads, we introduced the temporal dimension to user-defined cuts and the graph generation process. Each operator can be partitioned along the spatial dimension, the temporal dimension, or both. The fine-grain graph is generated with proper inter-layer edges, as well as intra-layer edges. Intra-layer edges ensure a proper temporal schedule and an efficient spatial schedule. The IDs of the computation tiles include spatial and temporal dimensions.

\subsubsection{Intra-layer mapping}

In this stage, energy and latency are estimated for each unique CT-core combination using Zigzag \cite{zigzag}. If a CT does not fit on a core, the external memory is added to the core during cost estimation. Zigzag explores intra-layer mapping by breaking down the mapping space to loop prime factors. These factors are permuted and partitioned into different memory levels for each operand. For each valid mapping (i.e. tiles fit into their allocated memories), latency and energy are estimated. The latency and energy models translate the temporal mapping into data movement rates and volumes and allocated memory sizes. The latency model can identify memory-bound and compute-bound schedules and estimate latency accordingly. Memory-bound schedules occur because of data movement rates that are larger than the available memory bandwidth. The energy model is based on the volume of data movement between different levels of memory and the number of operations. We configure Zigzag to choose the best cost estimates based solely on energy.

To support SNNs, neuron state behavior and time are incorporated into Zigzag. Neuron states are added as explicit operands and their behavior is modeled into the memory allocation and data movement models. As a result, their costs are reflected in the latency and energy models. On the other hand, time is added as an explicit dimension of workloads (e.g., SNN layers), which increases the size of the temporal mapping space. The neuron states are reused over different timesteps; however, there is a chronological dependency between neuron updates over time, due to the non-linear activation at the end of each timestep. In other words, evaluating a neuron (and updating its state) in one timestep requires evaluating it first for previous timesteps. This scheduling constraint limits and reduces the temporal mapping space, where input-relevant loops (e.g. input channels, kernel projection) need to be iterated before temporal loops can be iterated (and possibly batched together). Between successive temporal iterations, all iterated neurons (i.e., iterated output-relevant loops such as output channels and spatial dimensions) are activated, and their output spikes are generated.

\subsubsection{Inter-layer mapping}
\label{sec:schedule}
In this stage, computation tiles are scheduled according to layer-to-code allocations. For multi-core accelerators, different layer-to-core allocations are explored using a genetic algorithm which optimizes latency and energy. Latency and energy costs of one allocation are generated using an inter-core graph scheduler. 

The inter-layer scheduling order is defined according to the workload partitioning. If a layer is spatially partitioned, it is said to be in a layer-fused schedule. Additionally, a new scheduling dimension is introduced for SNNs and other recurrent networks; time (sequence). In contrast to layer fusion (Section \ref{sec:lf}), batching multiple timesteps enlarges intermediate features. However, it increases memory reuse (neuron state and weights) across time (\textit{temporal memory reuse}). Such effects are similar to input batching. In addition, neuron states are typically reset to zero at the beginning of an inference. Hence, batching all timesteps maximizes reuse of memory over time and may completely eliminate the off-chip data movement of neuron states under memory constraints. We call this a time-batched schedule (TB), in contrast to single-timestep schedule (ST) where only one timestep is scheduled at a time. Between TB and ST, we can batch just a few timesteps (e.g., 2T, 4T, 8T).

\begin{figure*}[!t]
    \centering
    \includegraphics[width=\textwidth]{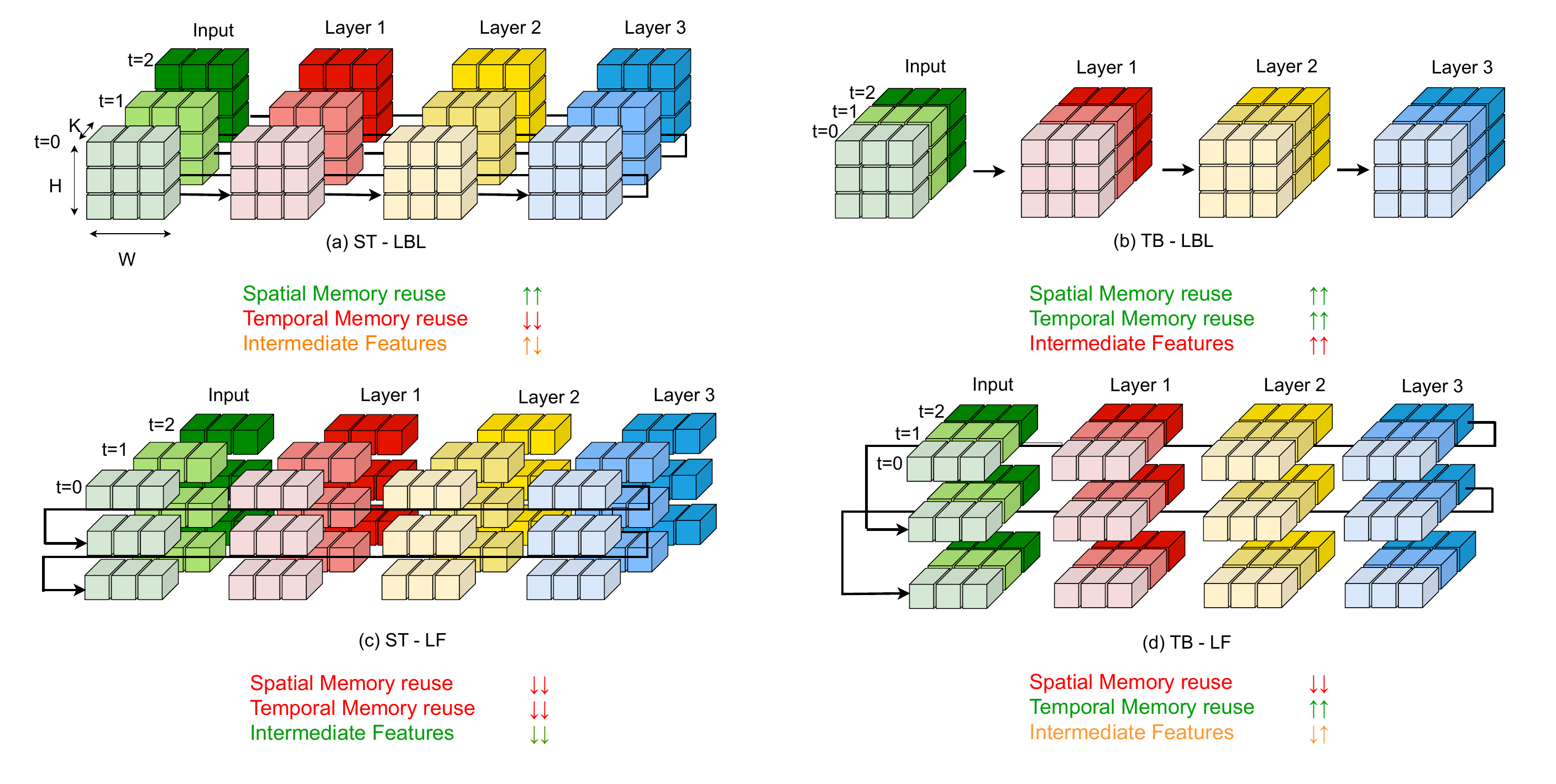}
    \caption{Illustration of different inter-layer schedules: (a) single-timestep layer-by-layer (ST-LBL) schedule, (b) time-batched layer-by-layer (TB-LBL) schedule, (c) single-timestep layer-fused (ST-LF) schedule, and (d) time-batched layer-fused (TB-LF) schedule for a 3-layer 3-timestep SNN block with 1D convolution connection (for simplicity). Some arrows are removed for readability. Each layer has output dimensions of size T x K x H x W, where K is number of channel, T is the number timesteps, and H and W are the height and width respectively. Time batching increases reuse of memory over time (temporal memory reuse) at the expense of larger intermediate feature sizes, while layer-fusion decreases reuse of memory in space (spatial memory reuse) but has smaller intermediate feature sizes.}
    \label{fig:schedules}
\end{figure*}

Temporal techniques can be applied orthogonal to spatial techniques (LBL and LF), resulting in different spatio-temporal schedules. In Figure \ref{fig:schedules}, we illustrate four combinations of spatio-temporal schedules which we use in our inter-layer schedule exploration: \textit{single-timestep layer-by-layer} (ST-LBL), \textit{time-batched layer-by-layer} (TB-LBL), \textit{single-timestep layer-fused} (ST-LF), and \textit{time-batched layer-fused} (TB-LF). Each schedule provides a different trade-off between spatial memory reuse, temporal memory reuse, and intermediate feature sizes. Batching $T$ timesteps \textit{increases} size of intermediate features and temporal memory reuse by $T$, while layer fusion \textit{decreases} size of intermediate features and spatial memory reuse, resulting in the different schedule properties highlighted in Figure \ref{fig:schedules}. Workload blocks with large feature sizes tend to profit from layer fusion, as it reduces the size of intermediate features, while workload blocks with large memory sizes (weights and/or neuron states) tend to profit from time batching, as it increases memory reuse across time. These four types of schedules can be applied separately to different blocks in an SNN, catering to different block characteristics (memory vs. features). In addition to these schedule, we can have different degrees of time batching resulting in different degrees of temporal reuse and intermediate feature sizes. The inter-layer schedule is generated according to user-defined cuts, as in Figure \ref{fig:schedules}.

During scheduling, a memory management unit and a communication management unit are used to orchestrate inter-layer schedule. The communication management unit manages inter-core data links and schedules inter-core and off-chip data transfers, while the memory management unit manages and allocates necessary memory space on each core. We assume a scratchpad memory structure. If a scheduled computation tile is too large to fit on its core, all tensors stored on the core's scratchpad memory (other than those required) are evicted, and the external memory data link is blocked to service the current computation tile. Otherwise, tensors are only evicted to free up scratchpad memory space if necessary or if they are no longer needed. The tensors to be evicted are ranked based on their size and priority, where priority is the number of times this tensor would be used in the future.



\subsection{Hybrid Schedule Exploration}
\label{sec:hb}

Given a workload with $N$ blocks, we would like to explore different possible schedules per block. In this exploration, we consider the spatio-temporal schedules in Figure \ref{fig:schedules}. An exhaustive search of this space results in $4^N$ possible schedules, which is not feasible for larger network sizes. Instead, we apply our knowledge regarding the trade-offs of these schedules. We rank blocks according to their feature sizes and order, and according to their memory size and order. We apply layer fusion and/or time batching to blocks in this ranking order. This results in $N+1$ possible ways (from $0$ blocks to $N$ blocks) to apply fusion or time batching, resulting in a total of $(N+1)^2$ possible schedules.

\subsection{Neuron State Optimization}
\label{sec:mem}

Neuron states enable learning patterns over time. They provide the network with the sequential memory necessary to learn (spatio-)temporal tasks. Deep learning models learn spatial features hierarchically, creating high-level semantic features in deeper layers \cite{10.1145/1553374.1553453}. Recent research in training networks on spatiotemporal tasks has shown that temporal learning in earlier blocks provides little to no gain and comes at a significant memory cost \cite{red,10204090,yik2024neurobench,9749022}. The best strategy is often to apply spatial feature extraction in earlier blocks and only learn sequences from higher-level features in later blocks. This reduces model complexity and memory footprint and avoids learning dynamics of low-level features which are usually unnecessary and less stable


Hence, we propose a neuron state optimization targeting spatio-temporal problems such as event-based computer vision. We propose to remove neuron state memory (i.e. forget them between timesteps) from early SNN blocks. This optimization significantly reduces the number of neuron states. For example, removing neuron states from the first 11 blocks out of the 50 blocks of a SEW-ResNet-152 \cite{sewresnet} network reduces its neuron states by approximately 40\%. We remove neuron states while preserving the spike-based activation and surrogate gradient to have minimal effect on the computational model and make a fair comparison. In STEMS, we model such layers as normal ANN layers with spike-based features.

%% file: setup.tex
\section{Experimental Setup} \label{sec:expsetup}

We define our hardware model and the two event-based vision benchmarks that we use in our experiments. In these experiments, we use STEMS and our hybrid schedule exploration as defined in Section \ref{sec:methods}.

\subsection{Hardware Architecture}

\begin{figure}[t]
    \centering
    \includegraphics[width=.6\columnwidth]{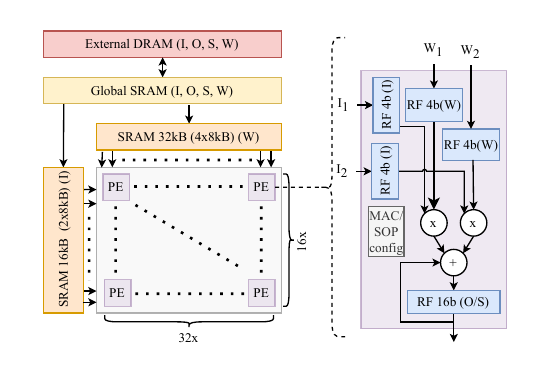}
    \caption{Meta VR Prototype \cite{Sumbul2022System-LevelAvatars} architecture diagram. (Left) Memory hierarchy and systolic array of PEs. (Right) PE details: reduction-of-2 MACs/SOPs. I: Input, W: Weight, O: Output, S: State.}
    \label{fig:meta_hw}
\end{figure}

For our hardware model, we opt for an output stationary architecture, as it seems more natural to maximize the stationarity of neuron states inside the stationary accumulators, similar to other neuromorphic architectures \cite{10132492, 9855834}. We adapt and modify the Meta VR prototype architecture \cite{Sumbul2022System-LevelAvatars} as our hardware model, shown in Figure \ref{fig:meta_hw}. The PE array is a 16x32 output-stationary systolic array. The accelerator contains two local buffers for inputs and weights and the PE array parallelizes 32 output channels and 16 output spatial positions. Additionally, a global SRAM is integrated on-chip, which we vary in size during our experiments. We ignore the non-linear activation function (e.g. LIF, ReLU) as it is performed much less often than input integration\cite{zigzag}.

Each PE has an accumulator, which holds either the partial output sum of an ANN layer or the membrane potential of an SNN layer. It accumulates two inputs in parallel, which can be spiking synaptic operation (SOP) or valued multiply-and-accumulate (MAC). We assume 4-bit weight and 4-b inputs (for non-spiking features) and adjust the local buffer sizes accordingly, as 4-bit bit-width is sufficient for accurate inference \cite{9996763, 10.1007/978-3-031-44207-0_34}. While the PEs use 16-bit accumulators to store the membrane potential or partial output sum, quantization research shows that 12 bits are sufficient for storing neurons in the global SRAM \cite{10.1007/978-3-031-44207-0_34}.


We estimate our hardware model in 22 nm FDX technology. By synthesizing and measuring the PE unit and the SRAM buffers, we estimate the model's energy parameters. DRAM energy is based on reported values in literature \cite{Vogelsang2010UnderstandingMemories, Gao2017TETRIS, Cavigelli2017Origami:Accelerator, Jouppi2021TenProduct, stream}. Table \ref{tab:mem_mac_energy} summarizes the energy cost of each component. We assume 128 KB SRAM banks for the global buffer as in \cite{Sumbul2022System-LevelAvatars}, regardless of its total size. However, global buffer energy has little effect on this work's results as most data movement is between the local buffers and PEs.


\input{mem_mac_energy}

\subsection{Benchmarks}


For our experiments, we use two SoTA SNN models for two event-based vision benchmarks. We use the Spike-Element-Wise (SEW-) ResNet-18\cite{sewresnet} model for the CIFAR10-DVS\cite{10.3389/fnins.2017.00309} dataset, and the hybrid RED-LIF\cite{yik2024neurobench} model for the Gen4 dataset\cite{red}. The SEW-ResNet CIFAR10-DVS model is fully LIF-based and is easy to train. Hence, it can be easily used for neuron state optimization, which requires multiple training sessions. Both models are trained using surrogate gradients and back-propagation through time\cite{8891809}.


\subsubsection{CIFAR10-DVS SEW-ResNet-18}

CIFAR10-DVS dataset \cite{10.3389/fnins.2017.00309} is a recording of CIFAR10 images using a DVS camera. The 128x128 DVS input is pre-processed into 16 timesteps with 2 input channels as in \cite{sewresnet}. SEW-ResNet is a recent architecture for deep residual learning in SNN \cite{sewresnet}, which applies a spike-element-wise (SEW) residual connection. The SEW-ResNet-18 model consists of 7 SEW blocks. Each SEW block consists of convolutional LIF layers, a SEW residual connection, and a 2x2 spike max-pooling. We use a novel SEW OR function (\textit{g}) instead of the SEW ADD function used in \cite{sewresnet}, to preserve the low precision of the spikes without losing accuracy compared to SEW ADD\cite{sewresnet}. 


Most of the SEW model's memory, features, and computations are in the first two blocks, with 80\% of neuron states and features in the first block, and 15\% in the second block. 

\subsubsection{Gen4 \textbf{R}ecurrent \textbf{E}vent-camera \textbf{D}etector (RED-LIF)} 

The Gen4 dataset is by far the largest event-based vision dataset currently available. It features hours of street recordings with millions of bounding boxes for cars, pedestrians, and bikers, recorded with a 1 Megapixel DVS camera \cite{red}. 

The original RED architecture was trained on the earlier Gen1 dataset. It is a \textit{hybrid model} which consists of two parts. First, it consists of feed-forward convolutional and squeeze-and-excitation layers to extract spatial features. Then, the high-level spatial features are passed through convolutional LSTM layers to learn spatial-temporal features.  LSTM layers are necessary for learning features over time. The RED model achieves a mean average precision (mAP) of 0.44 \cite{red}. Inspired by RED, RED-LIF  was developed with 3 feed-forward residual blocks followed by 5 convolution LIF layers. It achieves a mean average precision (mAP) of 0.29 on the Gen4 dataset\cite{yik2024neurobench}. Similarly to RED, the 640x360 DVS input stream is pre-processed into 6 channels and each training sample consists of 12 timesteps\cite{red}. 

The three feed-forward blocks contain most of the features and most of the computation, with 8.3 GMACs per timestep compared to 870 MSOPs per timestep in the LIF layers (ignoring sparsity). Most of the model's memory is in the LIF layers, with 2.6M weights and 1.3M neuron states compared to 0.4M weights in the feed-forward blocks. While the LIF layers have high sparsity (average 93\%), the feed-forward analogue blocks have an average sparsity rate of only 50\%. 



\subsection{Validation and scalability}

To verify STEMS, we created a cycle-accurate simulation of some of STEMS's generated schedules on our target hardware. The resulting data movement from our simulation matched the cost estimates generated from STEMS. To demonstrate STEM's scalability, we also performed our experiments on the SEW-ResNet-152 model. Results are reported in the appendix.

%% file: mem_mac_energy.tex
\begin{table}[t]
\centering
\caption{Energy cost per memory access and per compute.}
\label{tab:mem_mac_energy}
\begin{tabular}{@{}lrr@{}}
\toprule
Memory type & & R/W Energy (fJ/bit)  \\ \midrule
DRAM & & $11,000 / 12,000$  \\
SRAM (128kB/bank) & & $\approx 130 / 170$  \\
SRAM 32kB (4x8kB) & & $100 / 120$   \\
SRAM 16kB (2x8kB) & & $100 / 120$  \\ \midrule
 &  &  \\ \midrule
Operation & & PE Energy (fJ/OP) \\ \midrule
4-bit Multiply-and-accumulate (MAC) & & 160   \\
4-bit Synaptic Operation (SOP) & & 70  \\ \bottomrule
\end{tabular}
\end{table}

%% file: experiments.tex
\section{Experiments} 
\label{sec:experiments}

We perform three experiments. First, we highlight the effects of time batching on the reuse of neuron states. For that, we perform a time batching analysis on the SEW-ResNet-18 model, as it is a fully LIF-based model. Then, we explore our memory optimization from Subsection \ref{sec:mem} on the SEW-ResNet-18 model. Finally, we perform our hybrid schedule exploration on the RED-LIF model, the SEW-ResNet-18 model, and the optimized SEW-ResNet-18 model.

\subsection{Time batching analysis}

 In this experiment, we gradually apply time batching to the SEW-ResNet-18 model. We use a 1 MB global buffer, which is insufficient for storing all neuron states. Out of 16 timesteps (per sample), we batch 1T (ST), 2T, 4T, 8T, and 16T (TB).


\begin{figure*}[!t]
    \centering
    \includegraphics[width=\textwidth]{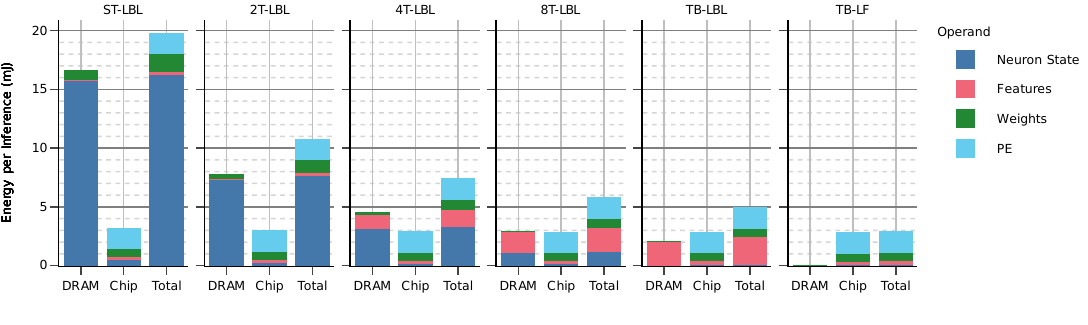}
    \caption{Effects of time batching on SEW-ResNet-18 model with 1MB on-chip memory; energy breakdown per inference.}
    \label{fig:sew7-tb}
\end{figure*}

As Figure \ref{fig:sew7-tb} shows, the more timesteps we batch, the less neuron states traffic to DRAM. When time is batched completely, neuron states never leak to DRAM, but are rather generated and discarded on-chip during an inference (16 timesteps). However, intermediate feature sizes grow larger the more we batch time. The spikes start to leak to DRAM after batching 4 timesteps, and it gets worse the more timesteps you batch. Applying a layer fusion schedule in combination with time batching can counteract the effect of enlarged intermediate features in time by shrinking them in space, at the cost of weight (spatial) reuse.


\subsection{SEW-ResNet-18 neuron state optimization}
\label{sec:memop}


We perform an ablation study on the neuron states of the SEW-ResNet-18 model. Following the approach of Section \ref{sec:mem}, we remove the hidden states from earlier blocks while preserving the same spike-based functionality. The results of our study are shown in Table \ref{tab:sew}. We define each model by the number of blocks with neuron states (SEW-X). 


\begin{table}[!t]
\caption{SEW-ResNet-18 hidden state memory ablation study.}
\label{tab:sew}
\centering
\begin{tabular}{|c||c|c|c|}
\hline
Model & Test Accuracy & \# Neuron States & \% Neuron States\\
\hline
SEW-7 & 73.5\% & 46M & 100\%\\
SEW-6 & 72.1\% & 8.6M & 20\%\\
\textbf{SEW-5} & \textbf{74.3\%} & 2.3M & 5\%\\
SEW-4 & 73\% & 750K & 2\%\\
SEW-3 & 73.2\% & 360K & 1\%\\
SEW-2 & 67.1\% & 61K & 0.1\% \\
SEW-1 & 67.4\% & 12K & 0.03\%\\
SEW-0 & 65.4\% & 0 & 0\%\\
\hline
\end{tabular}
\end{table}

By removing the neuron state from the first two blocks (SEW-5), we reduce the memory of the neuron state of the model by 95\% and slightly improve its test accuracy. Note that SEW-5 has similar structural properties as ANNs and RED-LIF where most features and computation are in earlier blocks (blocks 0 and 1), and most memory is in later blocks. While in the original model (SEW-7), most features, memory, and computation are in block 0 and 1. Assuming 12-bit states and 4-bit weights, the neuron states in SEW-7 require 5.5 MB of memory, while in SEW-5 require only 0.3 MB of memory, and weights require 0.5 MB. SEW-7 and SEW-5 models have the same average sparsity rates of 93\%. Except for the input layer, both models perform SOP operations. 

\subsection{Hybrid schedule exploration}
\label{sec:Shd}


We perform hybrid schedule exploration on the RED-LIF model, and the original (SEW-7) and optimized (SEW-5) SEW-ResNet-18 models, under different on-chip memory constraints. We choose stringent memory sizes (512 KB for RED-LIF and 128 KB for SEW-ResNet-18) to highlight the benefits of inter-layer optimizations in reducing memory requirements for efficient inference. We report here a few of our results and report the rest of the results in the appendix. 




\subsubsection{RED-LIF}
We present the results of hybrid schedule exploration for RED-LIF with a 512 KB on-chip global buffer. We explore time batching starting from the output block and layer fusion starting from the input block, since the feed-forward part contains most of the model's features, while the LIF layers contain most of the model's memory.

\begin{figure}[t]
    \centering
    \includegraphics[width=.7\columnwidth]
    {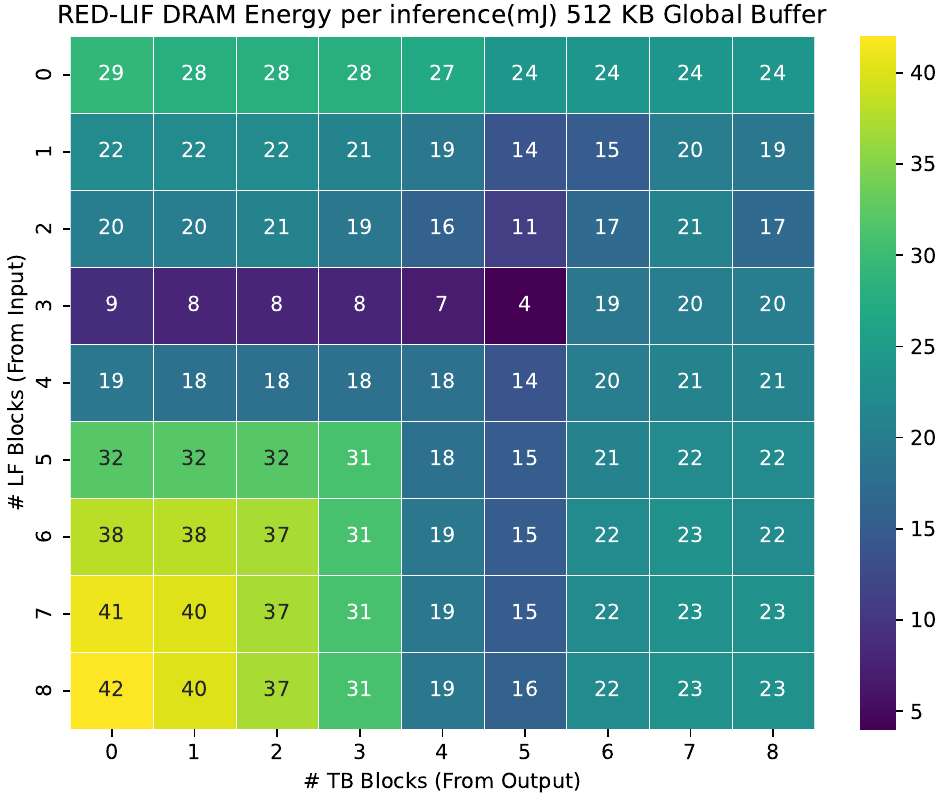}
    \caption{RED-LIF hybrid schedule exploration with 512 KB on-chip memory; DRAM energy per inference.}
    \label{fig:redlif-512}
\end{figure}

\begin{figure}[t]
    \centering
    \includegraphics[width=.7\columnwidth]
    {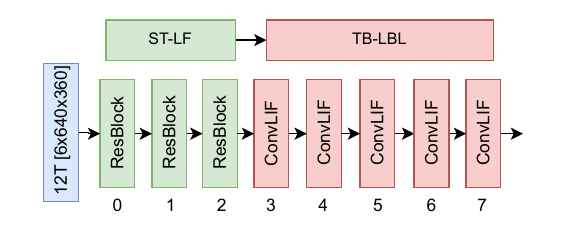}
    \caption{Optimal hybrid RED-LIF schedule with 512 KB on-chip memory.}
    \label{fig:red-best}
\end{figure}

Figure \ref{fig:redlif-512} shows the DRAM energy consumed per inference (12 timesteps) for all 81 possible schedules, where the horizontal axis represents the number of blocks that are time-batched (in this case, from the output side) and the vertical axis represents the number of blocks that are layer-fused (always from the input side). The best schedule deploys a single-timestep layer-fused (ST-LF) schedule for the feed-forward blocks and a time-batched layer-by-layer (TB-LBL) schedule for the LIF layers. Such results agree with the claim that blocks with more features favor schedules that minimize intermediate features, while blocks with more memory favor schedules that maximize memory re-use. The optimal hybrid schedule is illustrated in Figure \ref{fig:red-best}.

\begin{figure}[t]
    \centering
    \includegraphics[width=.7\columnwidth]{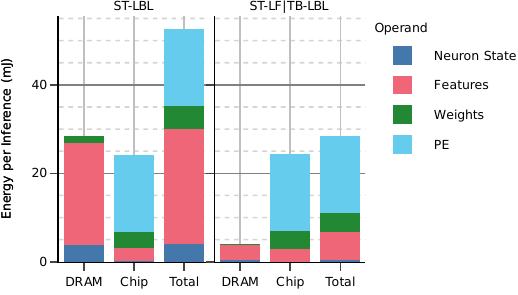}
    \caption{RED-LIF baseline schedule and optimal hybrid schedule with 512 KB on-chip memory; energy breakdown per inference.}
    \label{fig:red-sch}
\end{figure}

Figure \ref{fig:red-sch} shows a complete breakdown of the energy consumed by the baseline schedule (ST-LBL) and the optimal hybrid schedule (ST-LF$|$TB-LBL). Inter-layer optimizations reduced off-chip data movement by 7.3x and energy consumption by 1.9x, compared to the baseline schedule.


\subsubsection{SEW-ResNet-18}

\begin{figure*}[!t]
    \centering
    \includegraphics[width=\textwidth]{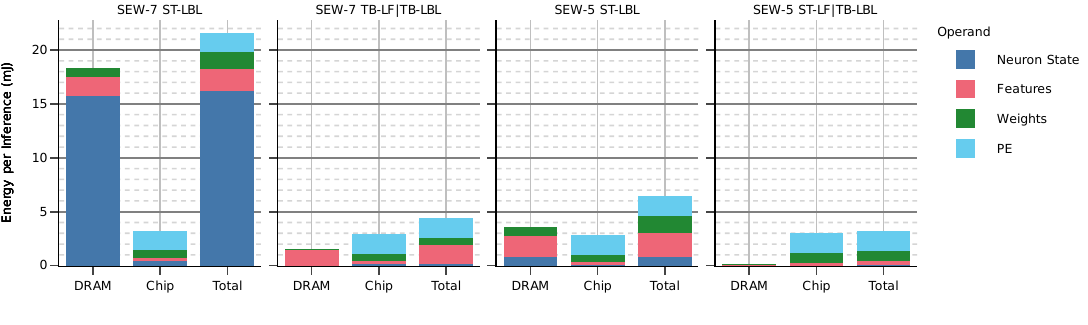}
    \caption{SEW-7 and SEW-5 baseline schedules and optimal hybrid schedules with 128 KB on-chip memory; energy breakdown per inference.}
    \label{fig:sew57-sch}
\end{figure*}

We present the results of the hybrid schedule exploration for the SEW-ResNet-18 with a 128 KB on-chip global buffer. For the original model (SEW-7), we explore both time batching and layer fusion starting from the input block, as earlier blocks have  large amounts of both features and memory due to their large LIF layers. For the optimized model (SEW-5), we explore time batching starting from the output block, and layer fusion starting from the input block, as it does not have LIF neurons in the early blocks. Figure \ref{fig:sew57-sch} shows a complete breakdown of the energy consumed per inference (16 timesteps) by the baseline schedules and the optimal hybrid schedules for SEW-7 and SEW-5. 

\begin{figure}[t]
    \centering
    \includegraphics[width=.7\columnwidth]
    {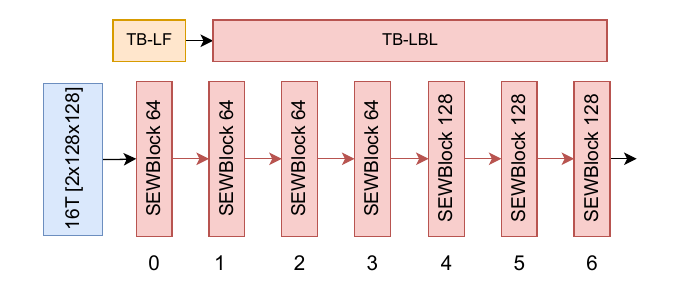}
    \caption{Optimal hybrid SEW-7 schedule with 128 KB on-chip memory.}
    \label{fig:sew7-best}
\end{figure}

For the SEW-7 model, the optimal hybrid schedule prioritizes time batching (temporal reuse of memory) throughout the workload, as shown in Figure \ref{fig:sew7-best}. However, earlier blocks that have more features and less weights prioritize layer fusion to reduce the size of intermediate features at the expense of (spatial) weight reuse, while later blocks that have more weights and fewer features prioritize layer-by-layer scheduling to maximize (spatial) weight reuse at the expense of larger intermediate features. Inter-layer optimizations reduced off-chip data movement by 12x, and energy consumption by 5x, compared to the baseline schedule.

\begin{figure}[t]
    \centering
    \includegraphics[width=.7\columnwidth]
    {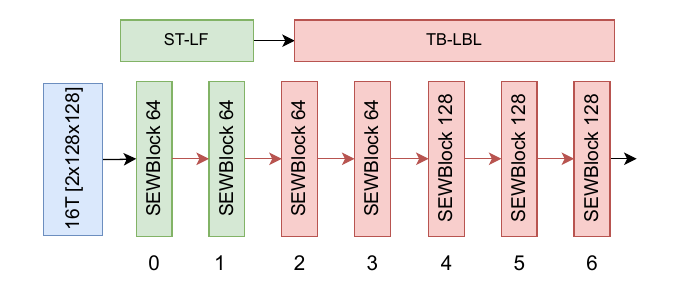}
    \caption{Optimal hybrid SEW-5 schedule with 128 KB on-chip memory.}
    \label{fig:sew5-best}
\end{figure}

For the SEW-5 model, the optimal hybrid schedule is similar to that of RED-LIF, as shown in Figure \ref{fig:sew5-best}. It deploys a single-timestep layer-fused schedule (ST-LF) for the first two optimized blocks, to limit the size of intermediate features (spikes). After removing the hidden state from these blocks, they no longer have a large memory. For later blocks that contain LIF neurons, it deploys a time-batched layer-by-layer schedule (TB-LBL) to maximize memory reuse in space and time, since these blocks have more memory (weights and hidden states) and fewer features (spikes). Inter-layer optimizations reduced SEW-5's off-chip data movement by 18x, and energy consumption by 2x, compared to the baseline. 

If we compare the two models' baseline schedules, SEW-5 has 5x less off-chip data movement and 4x less energy consumption than SEW-7. If we compare their best schedules, SEW-5 has 7.5x less off-chip data movement and 1.4x less energy consumption than SEW-7. The energy gains are diminished for the best schedules compared to the baseline schedules because DRAM is less dominant for the best schedules.

%% file: discussion.tex
\section{Discussion} \label{sec:discussion}

This study has led to some interesting insights on SNNs.

\subsection*{Proper scheduling can mitigate the effects of neuron state}

In our experiments, we have shown how spatial-temporal inter-layer mapping optimizations can avoid a lot of DRAM traffic. Under stringent memory capacity, a single-timestep layer-by-layer schedule causes an off-chip traffic of neuron state that is detrimental to energy efficiency (Figure \ref{fig:sew7-tb}). Inter-layer scheduling optimizations drastically reduce on-chip memory requirements for efficient scheduling.

\subsection*{Optimal schedules are often hybrid schedules}

We have seen in our experiments how optimal schedules are hybrid, where different blocks favor different scheduling strategies. Early blocks that have no hidden states favor a single-timestep layer-fused (ST-LF) schedule, as they have more features and less memory due to their large spatial dimensions and shallow channel dimension. Late blocks favor time-batched layer-by-layer (TB-LBL) schedules, as they have more memory and less features due to their small spatial dimensions and deep channel dimension. Early blocks in a full SNN model favor time-batch layer-fused (TB-LF) schedules, as they have more features and memory (neuron states). This is demonstrated in Figures \ref{fig:red-best}, \ref{fig:sew7-best}, and \ref{fig:sew5-best}.


\subsection*{Event-based vision SNN models do not need most neurons}

We have shown how most features and neuron states of a computer vision SNN model are in its earlier blocks. The neuron states in such blocks are expensive and have low accuracy returns. Hybrid models consisting of an ANN followed by an SNN are more suitable for event-based vision than fully recurrent models. This is due to the following reasons:

\subsubsection*{Memory benefits}

Not having hidden states in earlier blocks leads to significantly less neuron state overhead, as earlier blocks have large spatial dimensions, as shown in Table \ref{tab:sew}. Significantly reducing the number of neuron states may also allow them to easily remain in on-chip memory.

\subsubsection*{Structural benefits}

Time-batching (TB) and layer fusion (i.e. LF) act on different dimensions (temporal vs. spatial) for different reasons (memory reuse vs. reducing intermediate features). Having a hybrid structure leads to a more balanced workload structure where the two strategies (TB and LF) do not compete within the same critical blocks. In a hybrid model, earlier blocks have much more features than memory, leading to a layer-fused schedule, while later blocks have much more memory than features, leading to a time-batched schedule. The same applies to ANN models and input batching.

\subsubsection*{Algorithmic benefits}

Learning spatial features first and transforming raw data into higher-level spatial features make temporal learning arguably simpler\cite{red}. Our experiments on SEW-ResNet-18 agree with the literature that removing hidden states from earlier blocks has little effect on accuracy\cite{10204090}.

\subsection*{Neuron state optimization creates hybrid SNN models}
Reducing neuron states from earlier blocks, as done in Section \ref{sec:memop}, creates a structure similar to hybrid models. Even if the whole workload is spiking, the optimized blocks act as feed-forward layers as they do not have any hidden states. The optimized models have a distribution of features, hidden states, and weights similar to hybrid models. Additionally, our experiments show, under tight memory constraints, that SEW-5's best schedule is similar to RED-LIF's best schedule. Hence, optimizing neuron state memory, as in Section \ref{sec:mem}, brings the benefits of hybrid models to SNNs by improving their schedule, energy efficiency, and even accuracy. 

\section{Future Work}

We have the following suggestions for future work:




\subsection*{Improving on-chip energy}

This work focused on optimizing off-chip traffic with limited on-chip memory. Future work can look at improving the on-chip energy. For that, we have the following suggestions:
\begin{itemize}
    \item \textbf{Sparsity Utilization:} to reduce the compute cost of sparse SNN layers. 
    \item \textbf{Compute-in-memory:} to further reduce compute cost.
    \item \textbf{Heterogeneous accelerators:} to serve the different properties of hybrid models.
\end{itemize}
%




\subsection*{Code generation}

In this work, we used a simulation model for performance estimation only. Future work can be towards completing the loop with code generation for hardware. 

\subsection*{Exploring other recurrent models}

SNNs are self-recurrent models, where the hidden states of different neurons do not directly affect each other. Other recurrent models can have relationships between its hidden states. For example, in a convolution LSTM layer, the hidden states interact together through a convolution kernel. Hence, updating one state requires knowing other neighboring states. This prohibits time-batched layer-fused (TB-LF) schedules, due to the spatial dependency between states.

There are also other emerging transformer-like SNN networks \cite{yao2023spikedriven}. Such large networks have a completely different structure compared to image-based convolutional SNNs. Such networks can be explored in future work using STEMS.




%% file: conclusion.tex
\section{Conclusion} \label{sec:conclusion}

In this study, we investigated inter-layer mapping optimizations for event-based vision SNNs models using STEMS, our mapping exploration tool for SNNs. STEMS is the first tool to provide intra-layer and inter-layer mapping optimizations for SNNs and temporal workloads by extending recent tools to support SNNs and explore the spatio-temporal mapping space. Additionally, we enable our exploration by effectively trimming the inter-layer mapping space.

Our experiments show the benefits of layer fusion and time batching to the different stages of SNN models and hybrid SNN models; where time batching improves reuse of neuron states and weights (memory), while layer fusion helps reduce the size of intermediate spikes/features. Under stringent memory requirements, we have shown up to 12x reduction in off-chip traffic and up to 5x reduction in energy consumption. We have also shown how to turn an SNN model into a hybrid model, by minimally changing the neuron model to remove neuron states. We have shown how hybrid models can have a better structure for scheduling, better energy efficiency (1.4x), less memory footprint (20x less states), and even better task accuracy for event-based vision tasks.

\section*{Acknowledgment}
This work has been funded by the Dutch Organization for Scientific Research (NWO) as part of P16-25 eDL project 7.

%% file: appendix.tex


\null
\newpage
\begin{center}
\textbf{\large Supplementary information for STEMS: \\ Spatial-Temporal Mapping Tool For Spiking Neural Networks}
\end{center}

\begin{figure}[t]
    \centering
    \includegraphics[width=.7\columnwidth]
    {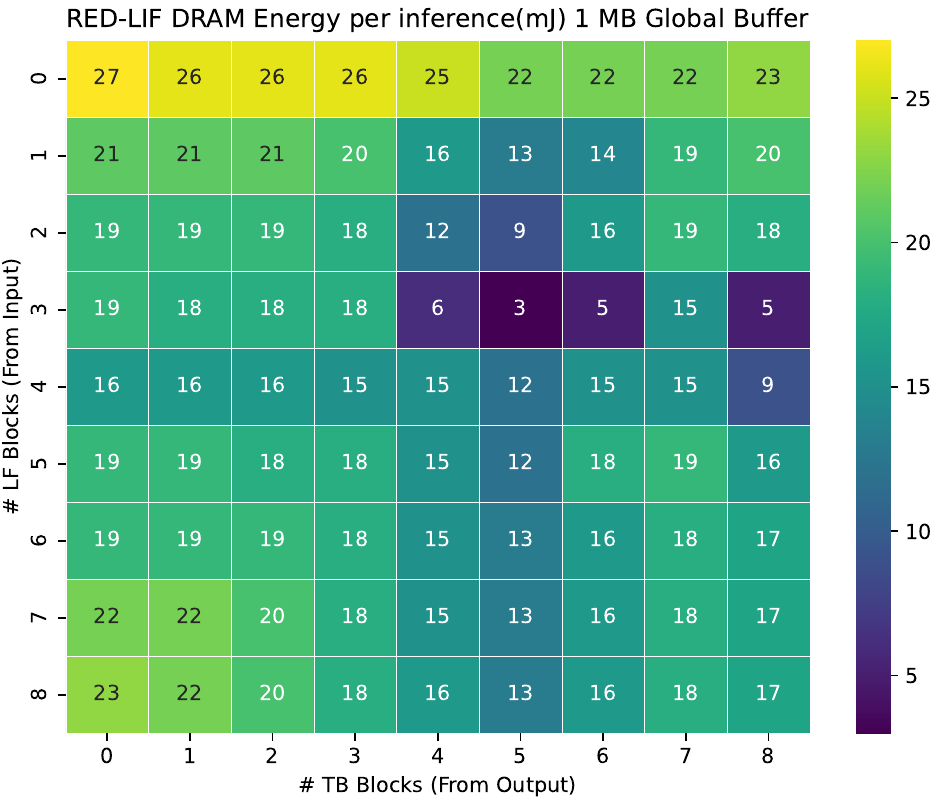}
    \caption{RED-LIF hybrid schedule exploration with 1 MB on-chip memory; DRAM energy per inference.}
    \label{fig:redlif-1m}
\end{figure}

\begin{figure}[t]
    \centering
    \includegraphics[width=.7\columnwidth]
    {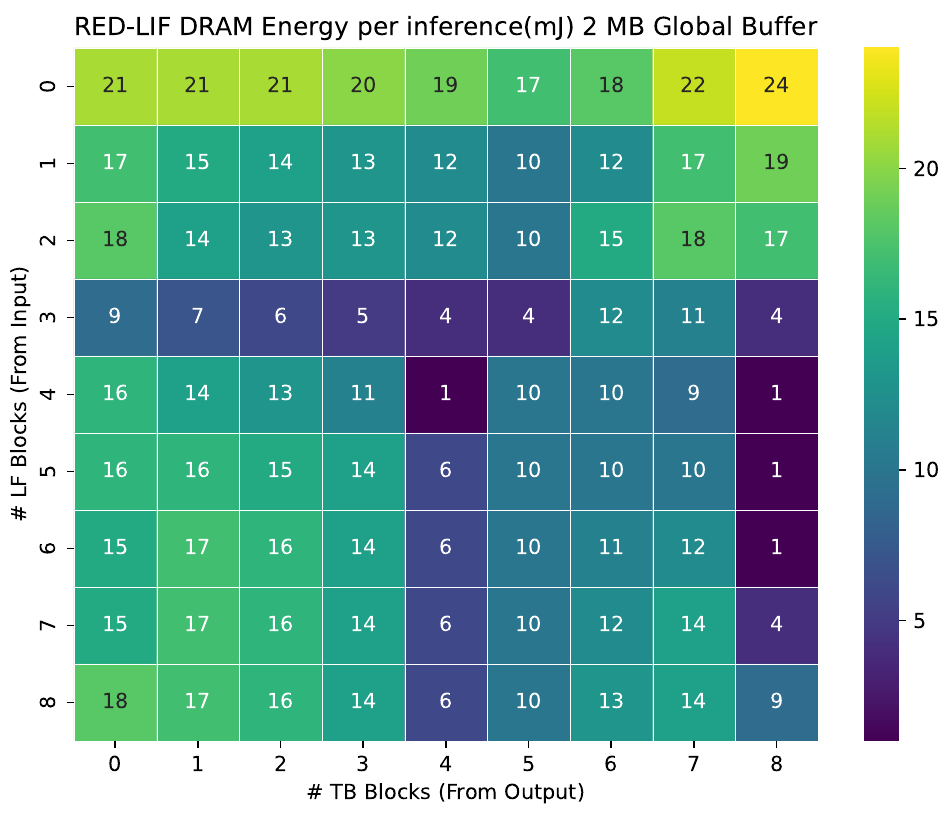}
    \caption{RED-LIF hybrid schedule exploration with 2 MB on-chip memory; DRAM energy per inference.}
    \label{fig:redlif-2m}
\end{figure}

\section*{Hybrid Schedule Exploration} \label{app:width_vs_accuracy}

We present here the remaining results of our hybrid schedule exploration. We perform exploratory studies for RED-LIF, SEW-7, and SEW-5 with different on-chip memory constraints. We present here the DRAM energy of all hybrid schedules. 

\begin{figure}[t]
    \centering
    \includegraphics[width=.7\columnwidth]
{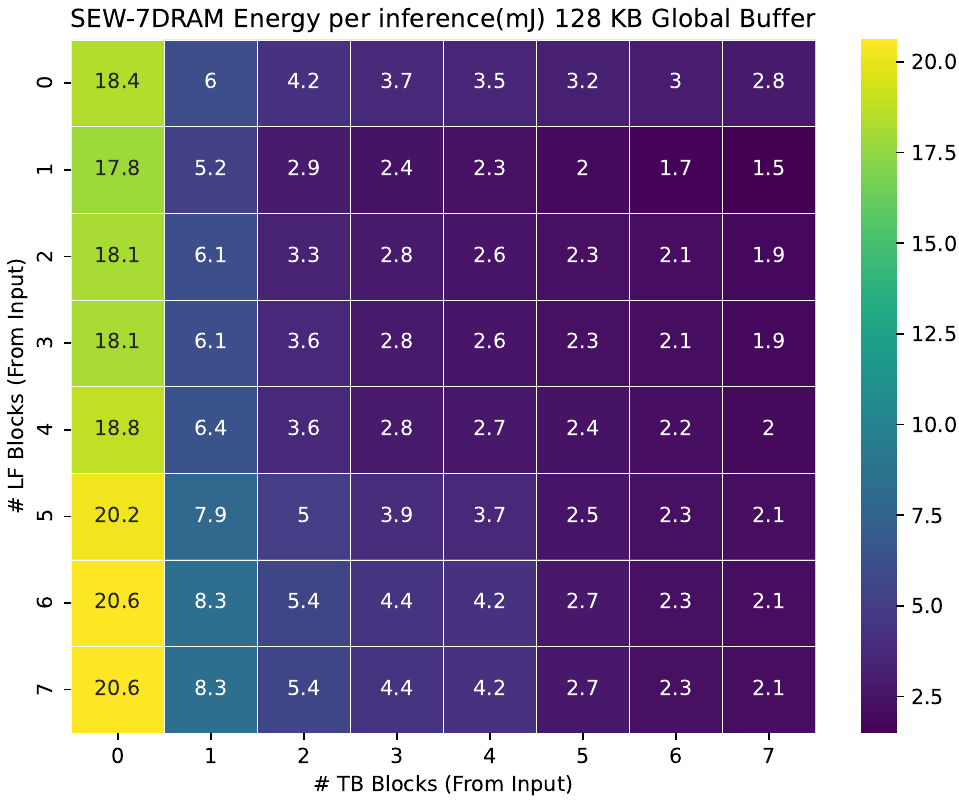}
    \caption{SEW-7 hybrid schedule exploration with 128 KB on-chip memory; DRAM energy per inference.}
    \label{fig:sew7-128}
\end{figure}

\begin{figure}[t]
    \centering
    \includegraphics[width=.7\columnwidth]
    {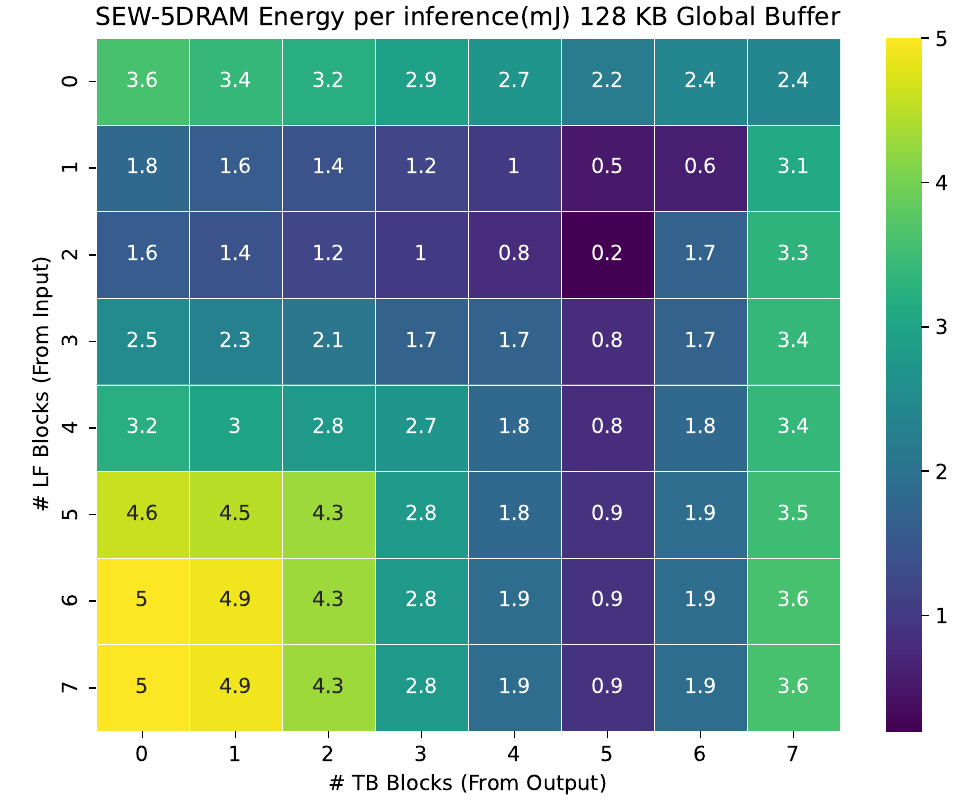}
    \caption{SEW-5 hybrid schedule exploration with 128 KB on-chip memory; DRAM energy per inference.}
    \label{fig:sew5-128}
\end{figure}

\begin{figure}[t]
    \centering
    \includegraphics[width=.7\columnwidth]
    {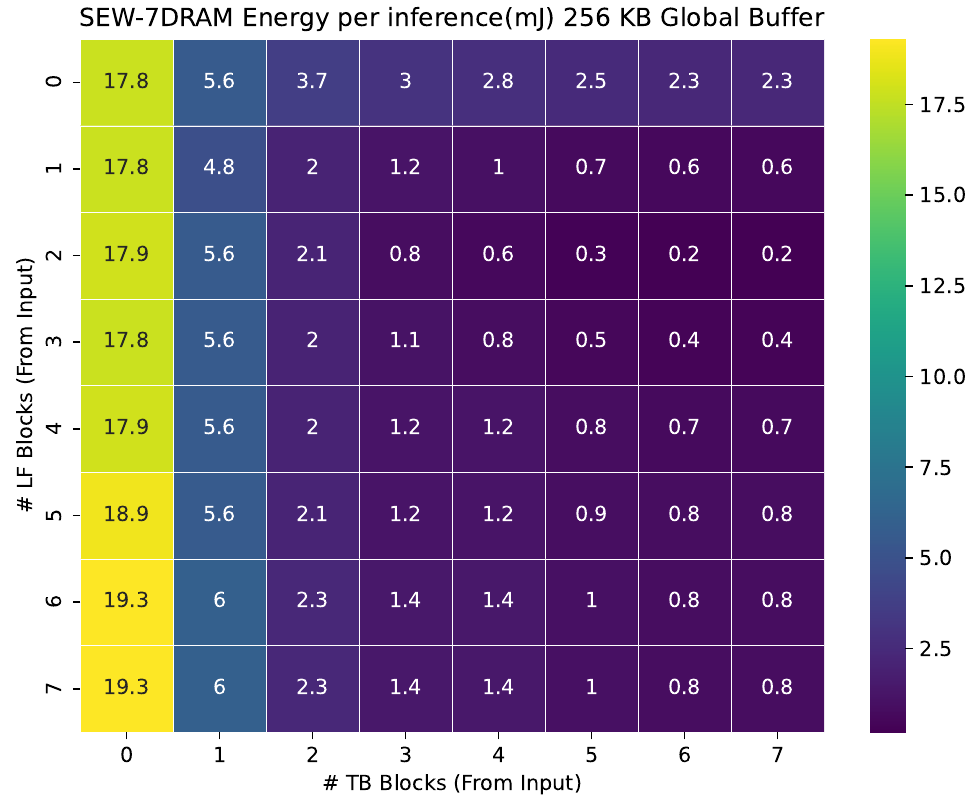}
    \caption{SEW-7 hybrid schedule exploration with 256 KB on-chip memory; DRAM energy per inference.}
    \label{fig:sew7-256}
\end{figure}

\begin{figure}[t]
    \centering
    \includegraphics[width=.7\columnwidth]
    {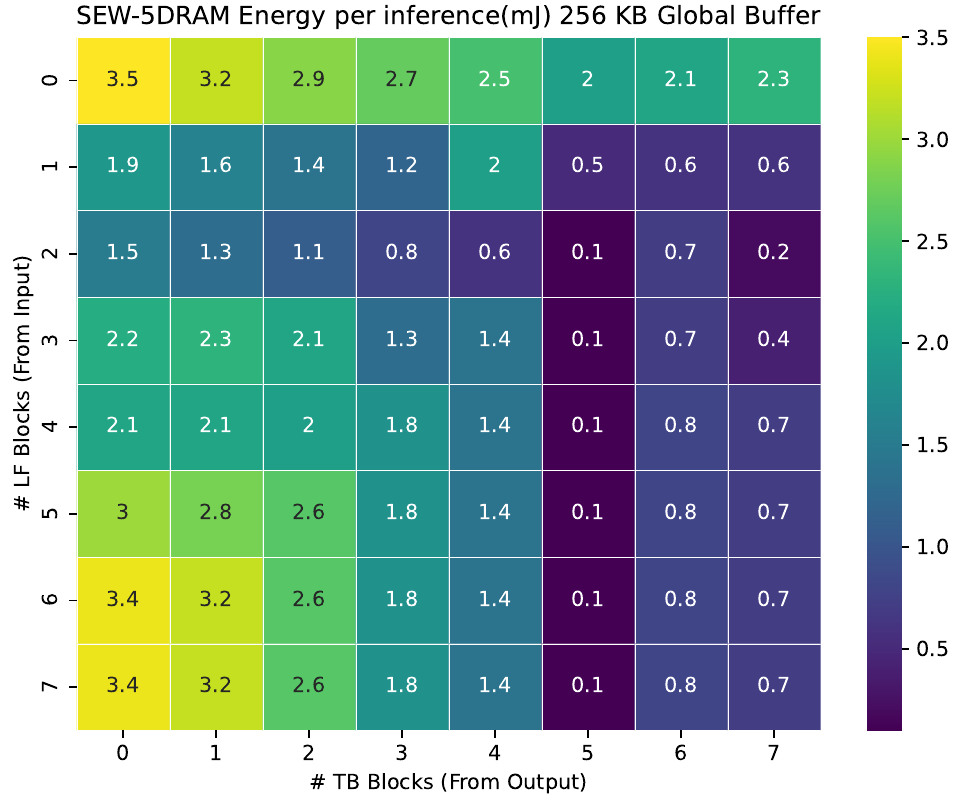}
    \caption{SEW-5 hybrid schedule exploration with 256 KB on-chip memory; DRAM energy per inference.}
    \label{fig:sew5-256}
\end{figure}

\begin{figure}[t]
    \centering
    \includegraphics[width=.7\columnwidth]{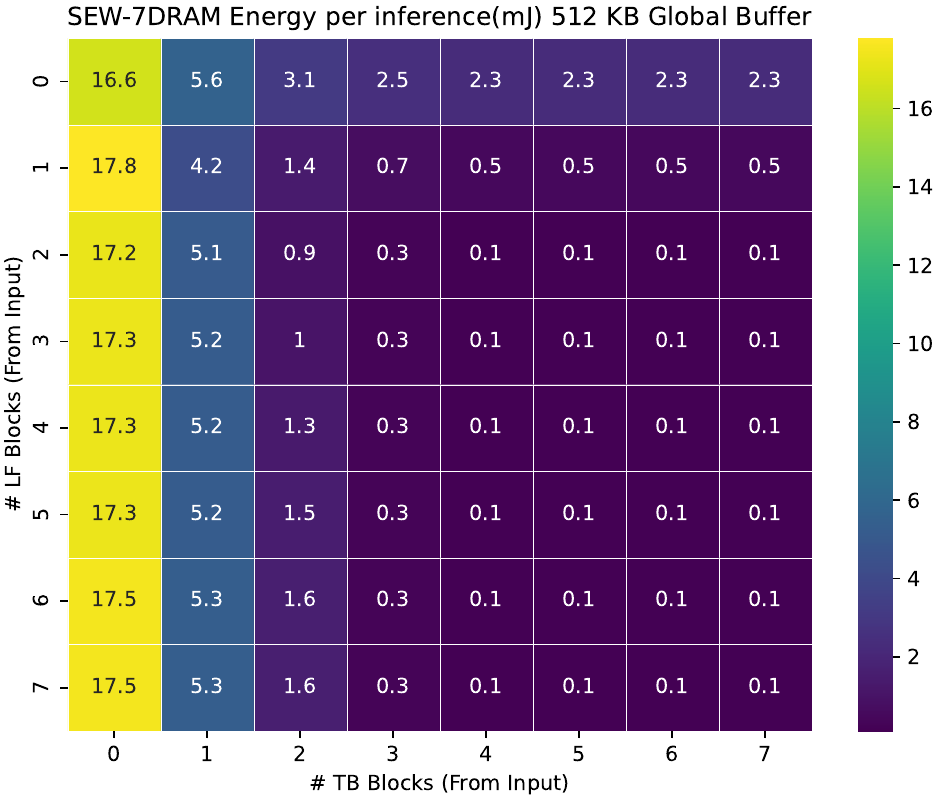}
    \caption{SEW-7 hybrid schedule exploration with 512 KB on-chip memory; DRAM energy per inference.}
    \label{fig:sew7-512}
\end{figure}

\begin{figure}[t]
    \centering
    \includegraphics[width=.7\columnwidth]{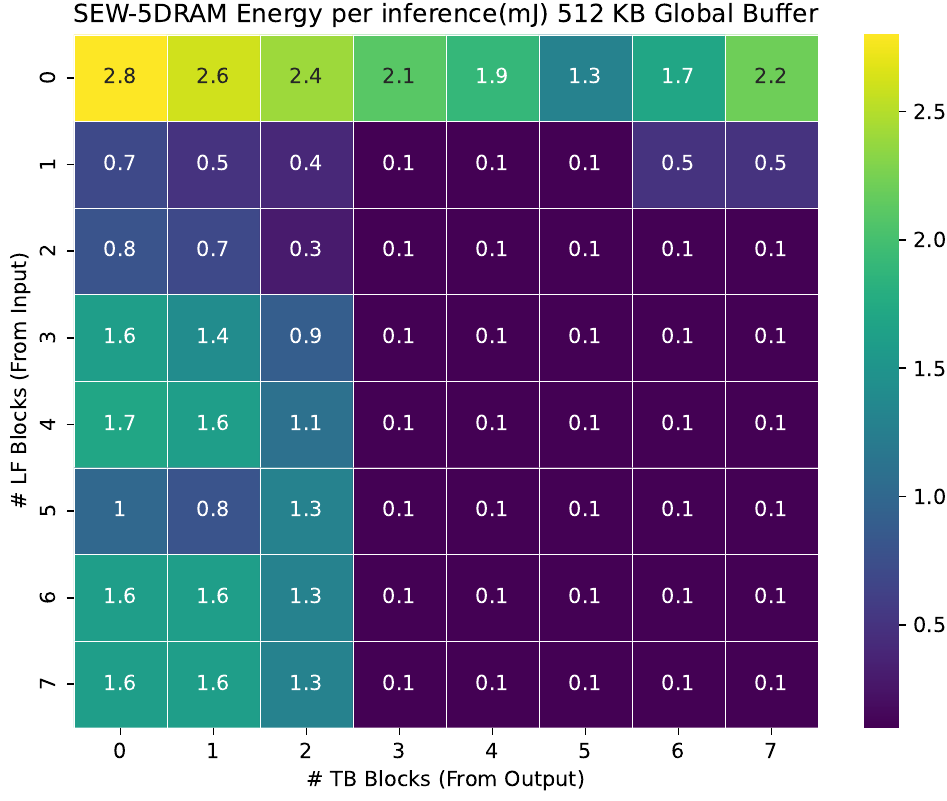}
    \caption{SEW-5 hybrid schedule exploration with 512 KB on-chip memory; DRAM energy per inference.}
    \label{fig:sew5-512}
\end{figure}

For RED-LIF, in addition to the exploration at 512 KB on-chip capacity (Figure \ref{fig:redlif-512}), we performed explorations at 1 MB (Figure \ref{fig:redlif-1m}) and 2 MB (Figure \ref{fig:redlif-2m}) on-chip capacity. For SEW-7 and SEW-5, we performed explorations at 128KB (Figures \ref{fig:sew7-128} and \ref{fig:sew5-128} respectively), 256 KB (Figures \ref{fig:sew7-256} and \ref{fig:sew5-256} respectively), and 512 KB (Figures \ref{fig:sew7-512} and \ref{fig:sew5-512} respectively) on-chip capacity.


Our results show that, under more relaxed on-chip memory, several schedules perform well. With enough on-chip memory, all schedules would perform equally well, with optimal DRAM traffic (1 mJ for RED-LIF, 0.1 mJ for SEW-ResNet). However, under tight memory constraints, only few schedules outperform the others. Such schedules are in line with our conclusions. Notice how the optimal schedule may vary depending on on-chip capacity.

\section{SEW-ResNet-152 ImageNet model}
\label{sec:apndx}

To demonstrate the scalability of STEMS, we explored the mapping space of a deeper model, SEW-ResNet-152. In \cite{sewresnet}, this model was used for ImageNet classification, where images were rate-encoded into 4 timesteps resulting in an input spike map of size 4x224x224. We assume the same input spike map in this experiment.

\subsection{Model structure}

The SEW-ResNet-152 model consists of 50 residual blocks organized by channel depth. It consists of 3 blocks with channel depth 64, 8 blocks with channel depth 128, 36 blocks with channel depth 256, and 3 blocks with channel depth 512. 

To reduce the number of experiments and have more readable exploration results, we coarsen the model into 7 hyperblocks as follows. The first hyperblock consists of the first 3 blocks with channel depth 64. Then, the second hyperblock consists of the 8 blocks with channel depth 128. Then, the 36 blocks with channel depth 256 are partitioned into 4 hyperblocks. Finally, the last 3 blocks with channel depth 512 represent the last hyperblock.

\subsection{Schedule exploration results}

We present the results of hybrid schedule exploration for SEW-ResNet-152 with 1 MB and 2 MB on-chip global buffer. We explore both time batching and layer fusion starting from the input block, as the earlier blocks contain large amounts of features and neuron states.

\begin{figure}[t]
    \centering
    \includegraphics[width=.7\columnwidth]{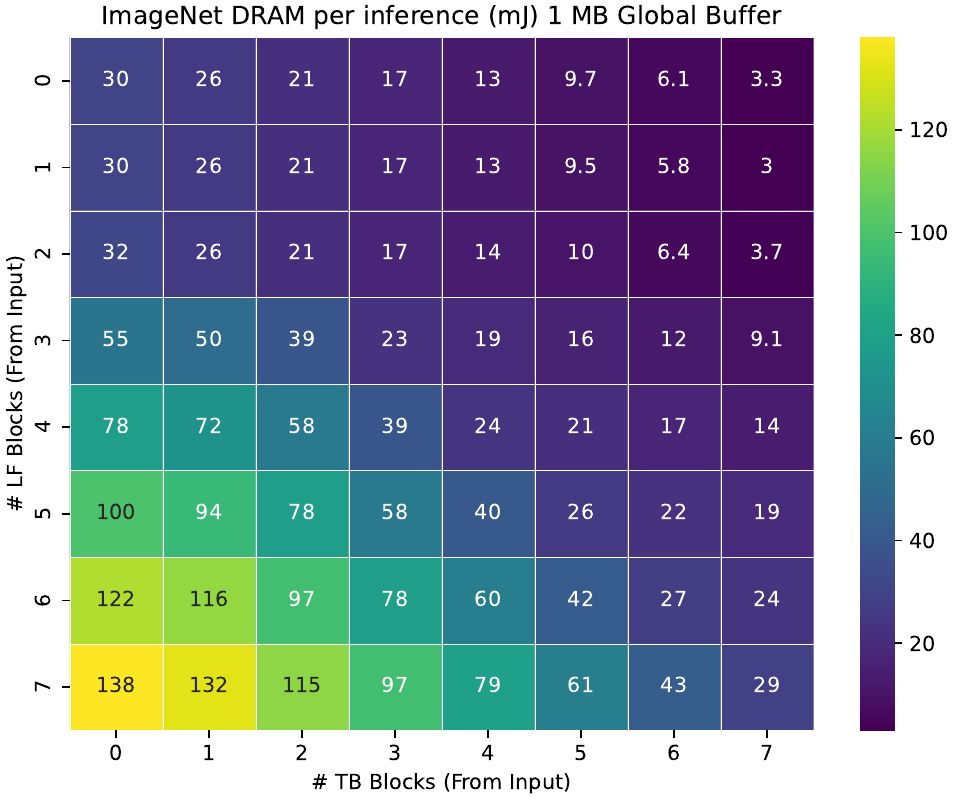}
    \caption{SEW-ResNet-152 hybrid schedule exploration with 1 MB on-chip memory; DRAM energy per inference.}
    \label{fig:sew152-1}
\end{figure}

\begin{figure}[t]
    \centering
    \includegraphics[width=.7\columnwidth]{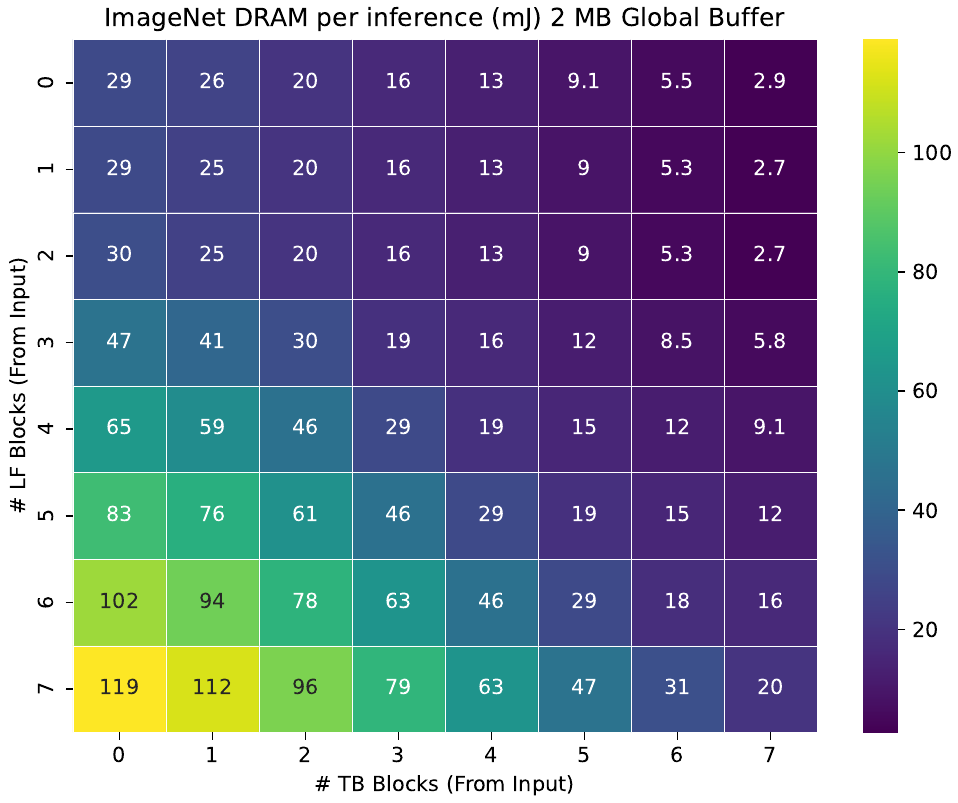}
    \caption{SEW-ResNet-152 hybrid schedule exploration with 2 MB on-chip memory; DRAM energy per inference.}
    \label{fig:sew152-2}
\end{figure}

Figures \ref{fig:sew152-1} and \ref{fig:sew152-2} show the DRAM energy consumed per inference (4 timesteps) for all 64 possible schedules mapped on the hardware architecture with 1 MB and 2 MB on-chip memory respectively, where the horizontal axis represents the number of blocks that are time-batched (from the input side) and the vertical axis represents the number of blocks that are layer-fused (from the input side). The optimal schedule deploys a time-batched layer-fused (TB-LF) schedule for the first hyperblock, and a time-batched layer-by-layer (TB-LBL) schedule for the rest of the network, for both hardware architectures. Such results are similar to the results of SEW-ResNet-18 hybrid schedule exploration (SEW-7), where earlier blocks favor TB-LF schedule and later blocks favor TB-LBL schedule. Applying this schedule results in 10x reduction in DRAM data traffic, compared to the baseline schedule, for both 1 MB and 2 MB on-chip memory architectures.

